\documentclass[conference]{IEEEtran}
\usepackage{fancyhdr}
\usepackage{lipsum} 
\usepackage{cite}
\usepackage{amsmath,amssymb,amsfonts}
\usepackage{algorithmic}
\usepackage{graphicx}
\usepackage{textcomp}
\usepackage{xcolor}
\usepackage[hidelinks]{hyperref}
\usepackage{makecell}
\usepackage{pgfplots}
\usepackage{multirow}
\usepackage{graphicx}
\def\BibTeX{{\rm B\kern-.05em{\sc i\kern-.025em b}\kern-.08em
    T\kern-.1667em\lower.7ex\hbox{E}\kern-.125emX}}
 
\fancypagestyle{plain}{
  \fancyhf{}
  \fancyfoot[C]{\thepage}%
}
\fancypagestyle{firstpage}{
  \fancyhf{}
  \fancyfoot[L]{\small © 2022 IEEE. Personal use of this material is permitted. Permission from IEEE must be obtained for all other uses, in any current or future media, including reprinting/republishing this material for advertising or promotional purposes, creating new collective works, for resale or redistribution to servers or lists, or reuse of any copyrighted component of this work in other works.}%
}  

\begin{document}

\title{Comparison of Missing Data Imputation Methods using the Framingham Heart study dataset
}

\author{\IEEEauthorblockN{Konstantinos Psychogyios, Loukas Ilias, and Dimitris Askounis}
\IEEEauthorblockA{\textit{Decision Support Systems Laboratory} \\ \textit{School of Electrical and Computer Engineering} \\
\textit{National Technical University of Athens}\\
15780 Athens, Greece \\
kwstaspsychogios@gmail.com, \{lilias,askous\}@epu.ntua.gr}
}

\maketitle

\begin{abstract}
Cardiovascular disease (CVD) is a class of diseases that involve the heart or blood vessels and according to World Health Organization is the leading cause of death worldwide. EHR data regarding this case, as well as medical cases in general,  contain missing values very frequently. The percentage of missingness may vary and is linked with instrument errors, manual data entry procedures, etc. Even though the missing rate is usually significant, in many cases the missing value imputation part is handled poorly either with case-deletion or with simple statistical approaches such as mode and median imputation. These methods are known to introduce significant bias, since they do not account for the relationships between the dataset's variables. Within the medical framework, many datasets consist of lab tests or patient medical tests, where these relationships are present and strong. To address these limitations, in this paper we test and modify state-of-the-art missing value imputation methods based on Generative Adversarial Networks (GANs) and Autoencoders. The evaluation is accomplished for both the tasks of data imputation and post-imputation prediction. Regarding the imputation task, we achieve improvements of 0.20, 7.00\%  in normalised Root Mean Squared Error (RMSE) and  Area Under the  Receiver Operating Characteristic Curve (AUROC) respectively. In terms of the post-imputation prediction task, our models outperform the standard approaches by 2.50\% in F1-score.
\end{abstract}


\section{Introduction}
\thispagestyle{firstpage}

Cardiovascular diseases (CVDs) cause the majority of deaths annually, taking  millions of  lives each year. CVD is a general term for conditions affecting the heart or blood vessels including coronary heart disease, cerebrovascular disease, heart valve disease and other conditions. Even though these types of diseases can be prevented, low and middle-income countries are vulnerable due to the lack of information and infrastructure  \cite{article13}.

Missing values is also a problem that occurs frequently within the clinical research framework. Missing data occurs when the values of the variables of interest are not measured or recorded for all subjects in the sample. Data can be missing for several reasons \cite{article36}, including: (i) patient refusal to respond to specific questions, e.g., patient does not report data on income; (ii) loss of patient to follow-up; (iii) investigator or mechanical error, e.g., sphygmomanometer failure; and (iv) physicians not ordering certain investigations for some patients, e.g., cholesterol test not ordered for some patients. When it comes to this type of issue the majority of researches choose to either delete these missing values or impute them using simple approaches such as mean, mode imputation or KNN \cite{article38}. The deletion choice is highly problematic since it  leads to a smaller dataset and a model that is not able to generalize well. Also, this approach  frequently produces results and errors that may be small for the complete subset of data but in reality are optimistic. The imputation alternatives lead to a complete dataset but are too simple and thus impute values that are unrealistic. Concerning CVD patient level data, there usually is a strong correlation between the corresponding variables, e.g., systolic and diastolic blood pressure, which should be incorporated to the missing value imputation model. This is something that univariate statistical approaches and simple regression algorithms fail to compute leading to inaccurate results \cite{article37}. These correlations of course are present in most medical datasets where tests have been conducted for the same patient, lab measurements have been carried out for a specific task etc.  

To address these limitations, in this paper we compare several methods of missing data imputation. Specifically, we propose two deep learning approaches based on Denoising Autoencoder (DAE) and Generative Adversarial Networks (GANs). The first approach is based on a DAE using kNN for pre-imputation. Using this model as baseline we implement various changes regarding both the architecture and the training process, which yield considerably more accurate results. In terms of the GAN approach, we also build upon the existing architecture making improvements concerning the specific case we study. To assess our models we use the Framingham heart study \cite{article16}. Finally, the proposed models are evaluated for both the imputation and post-imputation prediction tasks. We study the latter to explore whether the choice of more robust imputation methods will result in higher predictive performance or not. This is very important since prediction is usually the common goal of reasearchers and practitioners when applying machine learning techniques to EHR data.

Our main contributions can be summarized as follows:
\begin{itemize}
  \item We extend and improve existing missing value imputation models based on DAEs and GANs.
  \item We thoroughly evaluate several missing value imputation methods for various missing rates. 
  \item We prove that heart disease (and medical in general) EHR datasets require more sophisticated missing value imputation methods, especially when the missing rate is high.
  \item We show that robust deep learning methods can deal with both numerical and categorical data.
  \item We conduct post-imputation prediction and show that more accurate imputation leads to greater performance in the prediction task.
\end{itemize}

\section{Related Work}

\subsection{Missing Data Imputation in Healthcare}
Approaches towards missing value imputation employing EHR data vary in the literature. There is not a single solution that fits all cases and usually researches select algorithms that perform better to the specific task. The key reasons behind choosing an imputation technique are the mechanism of missingness, the gap length of missing data, etc.\cite{article35}.

The simplest way of dealing with this is deleting the records that contain at least one missing value and thus selecting a subset of the original dataset where there are no missing values \cite{article29,article30,article31,article32}. The authors in \cite{article2} studied the case of obesity prediction with EHR data using common machine learning models such as random forest and LSTMs. When it comes to missing values, they dropped rows with missing or corrupt values, e.g., implausible dates. At the same time, they dropped columns where more than 50\% of the entries were corrupt. Kwakye and Dadzie \cite{article11} studied the case of coronary heart disease using the Framingham heart study dataset, which is available at kaggle. Regarding the preprocessing steps, they chose to eliminate both missing and outlier data resulting in an undercomplete dataset.

Another popular but simple approach is the imputation based on mean, mode  or zero imputation where each missing value is imputed by zero \cite{article28}. 
Guo et al.\cite{article3} developed a deep learning approach for the problem of heart failure prediction using synthetic EHR data. In their analysis, they chose to discard features with more than 50\% of the entries missing and impute the rest using mean and most frequent for numerical and categorical features respectively. Gupta et al. \cite{article12} evaluated machine learning models for the problem of heart attack prediction using the Framingham heart study dataset and the Heart dataset for UCI Machine Learning Repository. When it comes to preprocessing and especially missing values, their approach was to impute using mean or median where the latter was preferred for features with skewed distributions.

Concerning categorical, missingness was dealt with the addition of an extra category for ‘missing’.
KNN and MLP are two methods that have also been used to address this issue. Jerez et al. \cite{article6} applied missing data imputation methods to a real world breast cancer dataset with overall 5.61\% percent of missingness. They utilized KNN, MLP, MICE, SOM, etc. algorithms for this problem. They found that the best performing method was KNN leading to a higher post imputation accuracy. 

Recent advances in deep learning have produced state-of-the art results modifying existing models to fit the missing value imputation framework. These advances, can be categorized as either discriminative or generative. Yoon et al. \cite{article8} modified the original GAN architecture and created GAIN. Results showed that it  this approach surpasses  robust imputation methods including an autoencoder based.

Regarding discriminative models, Aidos and Tom$\acute{a}$s \cite{article7} proposed an overcomplete DAE with KNN pre-imputation. Findings suggest that these methods outperform the standard ones and can handle high missing rates of up to 50\%. 

\subsection{Related Work Review Findings}
When it comes to works based on EHR data, we observe that many researchers do not give much emphasis on the missing value imputation part frequently choosing to drop these records. A very common approach also is to impute using simple methods such as mean, mode and KNN which are known to ignore complex relationships that define medical datasets. Concerning research using the framingham heart study we found no work that utilizes complex algorithms for the missing value imputation part even though the missingness in this dataset is significant. We also see that researchers do not evaluate imputation strategies by employing post-imputation prediction.

Therefore, our work differs significantly from the research works mentioned above, since we use state-of-the-art deep learning methods, make improvements regarding our case without loss of generality and additionaly test our imputation methods for the prediction task.

\section{Problem Statement}
\subsection{\textbf{Missing value imputation}}
The missing value imputation problem results if features of a dataset have unobserved values. Consider a random variable $X = (X_{1} , X_{2},..,X_{N})  \in X^{N}$  where $X$ represents the space to which each sample belongs to and has a distribution of $P(X)$. Consider also a mask vector $M = (M_{1} , M_{2},..,M_{N}) $  where each $M_{i}$ takes values in $\{0,1\}$ and $M_{i} = 1$ means the value is observed opposing to $M_{i} = 0$ which means the value is missing. Having $d$ instances of $X$ and $M$ we define a dataset $(X^{i},M^{i})$ for $i=0,1,..,d$. From this,  $(\hat{X}^{i},M^{i})$ is derived substituting each feature $j$ connected to a sample $i$ if $M_{i,j}=0$ with a pre-imputation value (possibly random noise). In such a case, given a model $IMP$ our goal is to create an imputed dataset $\tilde{X^{i}} = IMP(\hat{X^{i}},M^{i})$  for $i=0,1,..,d$. Each imputed sample should be generated based on $P(X|\tilde{X}=\tilde{X^{i}})$ since we want our imputed data to follow the original dataset's distribution.\\
The result is a new dataset $\bar{X}$ where for each sample $i$ we have :
$$\bar{X^{i}} = X^{i}\odot M^{i} + (1-M^{i})\odot\tilde{X^{i}}$$

\subsection{\textbf{Post-imputation prediction}}
Applying different missing value imputation algorithms $A_{i}$ for $i=0,1,..S$ to the original dataset $X$ results in $S$ different datasets $\bar{D_{1}},\bar{D_{2}},..,\bar{D_{S}}$. For each of these datasets we use a standard predictive method $P$ to predict the patient heart disease status (binary classification).

\section{Dataset}

We use the publicly available Framingham heart dataset \cite{article16}. The Framingham Heart Study is a longitudinal cardiovascular cohort study consisting of medical, lab, and questionnaire events on 4.434 participants. Originally, the dataset consists of 39 variables but we only used 15 for the purpose of this study as seen in Table \ref{tab:Table1}. This subset has 8 numerical and 7 categorical where the latter are all binary. The dataset is also not complete with feature missingness varying from 0 to 13\%.

\renewcommand\theadalign{bc}
\renewcommand\theadfont{\bfseries}
\renewcommand\theadgape{\Gape[1pt]}
\renewcommand\cellgape{\Gape[4pt]}

\begin{table}[h!]
\centering
\caption{Framingham heart study dataset.}
\label{tab:Table1}
\begin{tabular}{||c c c c||} 
\hline
 Feature & Description & Type & Missing  \\ [0.5ex] 
 \hline
 Sex & Male, Female & Categorical & 0 \\
 Totchol & \makecell{Serum Total \\Cholesterol \\(mg/dL) } & Numerical  & 409 \\
 Age & \makecell{Age in \\years } & Numerical& 0   \\
 SysBP &  \makecell{Systolic Blood  \\Pressure \\(mmHg) }& Numerical & 0\\
 Cursmoke &  \makecell{Current  smoking  \\at exam } & Categorical & 0  \\
 Cigpday &  \makecell{Number of  \\cigarettes smoked   \\each day} & Numerical& 79\\
 Bmi &  \makecell{Serum Total \\Cholesterol \\(mg/dL) } & Numerical& 52 \\
 Diabetes &  \makecell{Is Diabetic} & Categorical & 0 \\
 Bpmeds &  \makecell{Use of Anti  \\-hyp medication  \\at exam }& Categorical & 52 \\
 Heartrate &  \makecell{Heart rate \\beats/min} & Numerical& 6 \\
 Glucose &  \makecell{Casual serum  \\glucose (mg/dL)  } & Numerical & 1440 \\
 Prevhyp &  \makecell{Prevalent \\hypertention } & Categorical & 0 \\
 Prevstrk &  \makecell{Prevalent Stroke } & Categorical & 0 \\
 DiaBP &  \makecell{Diastolic Blood  \\Pressure \\(mmHg) } & Numerical& 0  \\
 CVD &  \makecell{Cardiovascular \\disease} & Categorical & 0  \\ [1ex] 
 \hline
 
 \end{tabular}

\end{table}

\section{Methods for Missing Data Imputation} \label{methods}

\paragraph{\textbf{Simple}}
This model is the simple statistical approach of mode, mean imputation. We impute categorical missing values using the most frequent class and numerical variables using the mean obtained by the corresponding column. 

\paragraph{\textbf{k Nearest Neighbors (KNN)}}
We employ KNN which imputes missing data considering the distance between the samples vectors in the dataset's space. For each feature missing, it considers the K, which is equal to 5, closest samples using euclidean distance that have this feature observed and averages their values.

\paragraph{\textbf{MissForest}}
We exploit the method proposed by \cite{article39}.

\paragraph{\textbf{Neighborhood aware autoencoder (NAA)}}
This autoencoder model was proposed by \cite{article7}, named neighborhood aware autoencoder (NAA). In this work, an overcomplete DAE was used for the problem of missing value imputation where the pre-imputation was done with KNN. The pre-imputation part is conducted for the whole dataset before model training using $k=5$. 

\paragraph{\textbf{Improved neighborhood aware autoencoder (I-NAA)}} \label{I-NAA}

Based on the NAA approach, we make some improvements. Firstly, we empirically choose an undercomplete architecture with half the input size for the encoder's output.\\
Regarding the training process of autoencoders, consider a copy $D_{copy}$ of the original complete dataset $D$. First we introduce missingness to $D_{copy}$ and then replace these missing values with noise or in our case with KNN imputation resulting in $D_{imputed}$. Then  batches of $D_{imputed}$ are fed iteratively to the network which must learn to map them to the corresponding batches of $D$. With constraints applied to the network the result of this process is that the autoencoder learns information about the relationships between the features.\\
However, if for the whole training process the train pre-imputed values of $D_{imputed}$ remain the same, the model may learn a mapping from these specific values to the actual ones and not the relationships between the features that define the dataset. Since our goal is both the latter and consideration of the local neighborhood, we change the pre-imputed values every \textit{N} epochs, where \textit{N} is empirically chosen. Specifically \textit{N} is equal to 10. Every 10 epochs, we still use KNN for pre-imputation but we change the value of k (closest neighbors) to one that hasn't been used before and is within certain $[B1,B2]$ bounds. We also create a custom loss function tailored to our dataset, namely RMSE for the numerical data and binary cross-entropy for the categorical.

\paragraph{\textbf{GAIN}}
This model was proposed by \cite{article8} and is based on the original GAN architecture. 

\paragraph{\textbf{Improved Gain (I-GAIN)}}
Based on GAIN approach, we propose some changes. Firstly, we add batch normalization both to the generator and the discriminator. Secondly, similar to the case of the autoencoder we pre-impute with KNN algorithm using different values of k (number of neighbors) every N epochs instead of random noise. We also utilize the same custom loss function as described in \textbf{\textit{I-NAA}}. In addition, the authors in \cite{article8} proposed a simple 3 layer architecture for the Generator part, where each layer had the same number of units. We replace this structure to one with 5 dense layers with an undercomplete autoencoder architecture since this model has proven to be effective in learning the dataset's distribution.

\section{Missing value imputation}
\paragraph{\textbf{Experimental Setup}}

To evaluate our methods we first select a subset of the original dataset that has no missing values. From the original 11627 records, 9310  remain. We then introduce artificial missingness to this dataset for comparison against the complete one. This step is performed univariately for the whole dataset, i.e., for each feature we remove $X \%$ of the values in a univariate manner, where $X$ can be 10, 20, 30, 40, and 50. Thus, we end up with a dataset $D_{imputed}$ which has  $X \%$ missingness for each feature. The mechanism behind the missing values introduction is MCAR \cite{article24}. To evaluate these models, we employ 5-fold cross validation and average the results between the 5 hold-out folds. The above procedure is repeated 10 times and the results are averaged. In total we train and evaluate each model $5\times10 = 50$ times.

\paragraph{\textbf{Evaluation Metrics}}
The metrics we choose are RMSE for the continuous features and Area Under the  Receiver Operating Characteristic Curve (AUROC) for the categorical. We use AUROC, since many categorical features are imbalanced, e.g., Diabetes has 8918 negative and 392 positive, and the former metric utilizes the probability of predictions. 

\paragraph{\textbf{Results}}
The results of our proposed methods mentioned in Section \ref{methods} are illustrated in Fig.~\ref{tab:Fig1} for the numerical features and in Fig.~\ref{tab:Fig2} for the categorical features.

Regarding the numerical features (Fig.~\ref{tab:Fig1}), we observe that our introduced methods, namely I-NAA and I-GAIN, are the best performing approaches achieving the lowest RMSE scores for all  missing rates. More specifically, I-NAA performs better than NAA for all missing rates, with the greatest difference observed at the missing rate of 10\% accounting for 0.04. It is also noted that I-GAIN improves the performance of GAIN in terms of all  missing rates achieving up to 0.06 lower RMSE for 30\% missingness. We can also see that the proposed deep learning approaches outperform the standard ones, i.e., Simple and KNN, for all missing rates. We finally notice that MissForest obtains better performance than Simple and KNN, while it achieves lower performance than I-NAA, I-GAIN, and NAA.

In terms of the categorical features (Fig.~\ref{tab:Fig2}), we observe that our introduced models, namely I-NAA and I-GAIN, yield the best AUROC results. Specifically, I-NAA outperforms NAA by 1.57-4.50\%, while I-GAIN improves the performance of GAIN by 1.30-6.00\%. We also notice that the deterioration of these methods with increasing high rate is slow and the performance is acceptable even for 50\% missingness. We also observe that MissForest performs better than GAIN in most missing rates, but it performs worse than all the other deep learning approaches. In addition, we note that the standard approaches, i.e., Simple and KNN, perform poorly. To be more precise, the performance of KNN ranges from 55.50\% to 73.50\%, while the Simple model yields a constant AUROC of 50.00\%, since it predicts only one class.

\begin{figure}
\centering
\begin{tikzpicture}
  \begin{axis}[ 
  width=\linewidth,
  line width=0.5,
  grid=major, 
  tick label style={font=\normalsize},
  legend style={nodes={scale=0.4, transform shape}},
  label style={font=\normalsize},
  legend image post style={mark=*},
  grid style={white},
  xlabel={Missing rate (\%)},
  ylabel={RMSE},
   y tick label style={
    /pgf/number format/.cd,
    fixed,
    fixed zerofill,
    precision=4
 },
legend style={at={(0.25,0.97)}, anchor=north east,  draw=none, fill=none},
  ]
    \addplot[blue,mark=*] coordinates
      {(10,0.112934) (20, 0.112168) (30,0.115363) (40,0.112389 ) (50,0.1119549)};
      \addlegendentry{Simple}
      
      \addplot[red,mark=*] coordinates
      {(10,0.1029231) (20,0.106913) (30,0.114294) (40,0.116217499) (50,0.117614)};
      \addlegendentry{KNN}
      
     \addplot[orange,mark=*] coordinates
      {(10,0.101812) (20,0.10644) (30,0.106608) (40,0.108748) (50,0.11083 )};
      \addlegendentry{MissForest} 
      
    \addplot[brown,mark=*] coordinates
      {(10,0.1015) (20,0.1038) (30,0.1051) (40,0.1051) (50,0.1065)};
      \addlegendentry{NAA}

   \addplot[black,mark=*] coordinates
      {(10,0.0976029) (20,0.1034) (30,0.104113) (40,0.1046567 ) (50,0.1063034)};
      \addlegendentry{I-NAA}
      
    \addplot[purple,mark=*] coordinates
      {(10,0.102812 ) (20,0.10544) (30,0.108381) (40,0.108577 ) (50,0.112024)};
      \addlegendentry{GAIN}
      
    \addplot[green,mark=*] coordinates
      {(10,0.1009816) (20,0.102833) (30,0.102833) (40,0.107471) (50,0.107024 )};
      \addlegendentry{I-GAIN}

  \end{axis}
\end{tikzpicture}
\caption{\label{tab:Fig1}RMSE comparison for various missing rates.}
\end{figure}
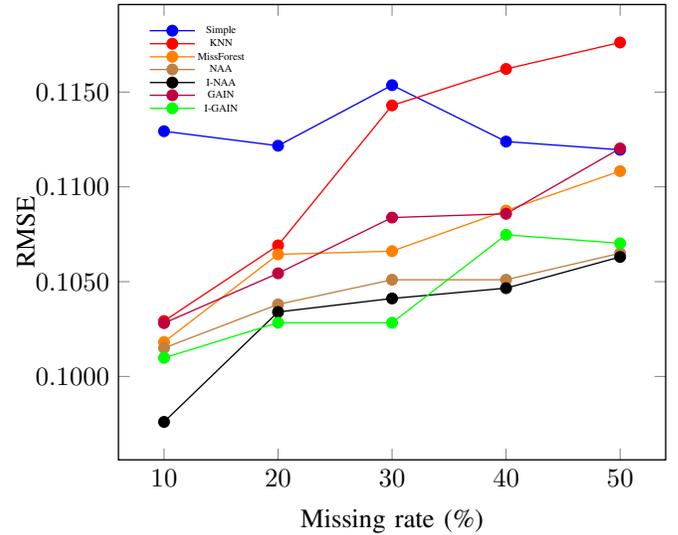

\begin{figure}
\centering
\begin{tikzpicture}
  \begin{axis}[ 
  width=\linewidth,
  line width=0.5,
  grid=major, 
  tick label style={font=\normalsize},
  legend style={nodes={scale=0.4, transform shape}},
  label style={font=\normalsize},
  legend image post style={mark=*},
  grid style={white},
  xlabel={Missing rate (\%)},
  ylabel={AUROC (\%) },
   y tick label style={
    /pgf/number format/.cd,
    fixed,
    fixed zerofill,
    precision=2
 },
legend style={at={(0.97,0.97)}, anchor=north east,  draw=none, fill=none},
  ]
    \addplot[blue,mark=*] coordinates
      {(10,50) (20, 50) (30,50) (40,50 ) (50,50)};
      \addlegendentry{Simple}
      
      \addplot[red,mark=*] coordinates
      {(10,73.5188) (20,67.7298) (30,60.85226) (40,56.878) (50,55.33)};
      \addlegendentry{KNN}
      
       \addplot[orange,mark=*] coordinates
      {(10,75.374) (20,73.424) (30,72.1) (40,70.43) (50,65.512 )};
      \addlegendentry{MissForest}
      
    \addplot[brown,mark=*] coordinates
      {(10,75.600) (20,73.030) (30,72.72) (40,71.405) (50,68.493)};
      \addlegendentry{NAA}

  \addplot[black,mark=*] coordinates
      {(10,80.125) (20,74.5704) (30,74.7726) (40,73.5363 ) (50,71.544)};
      \addlegendentry{I-NAA}
      
    \addplot[purple,mark=*] coordinates
      {(10,76.072 ) (20,73.299) (30,69.30) (40,69.00 ) (50,66.204)};
      \addlegendentry{GAIN}
      
    \addplot[green,mark=*] coordinates
      {(10,77.374) (20,75.424) (30,75.3) (40,73.78) (50,69.702 )};
      \addlegendentry{I-GAIN}

  \end{axis}
\end{tikzpicture}
\caption{\label{tab:Fig2}AUROC comparison for various missing rates.}
\end{figure}
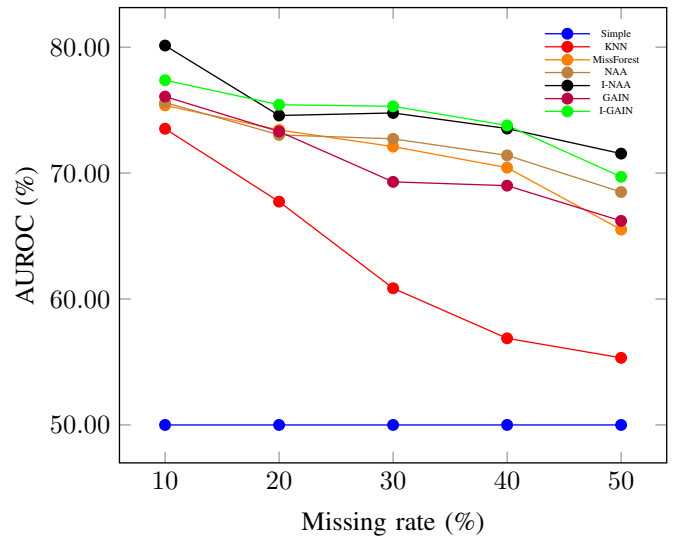

\section{Post-imputation Prediction}

\paragraph{\textbf{Experimental Setup}}
To further evaluate our models we conduct post-imputation prediction to see if better AUROC and RMSE scores also result in better performance for the prediction step. To do this for every model we save the complete, imputed dataset and perform a simple Random Forest to predict the CVD variable using 5-Fold cross validation and SMOTE for each fold's training set (since the dataset is highly unbalanced).

\paragraph{\textbf{Evaluation Metrics}}
For the post-imputation prediction task we use the F1-score.

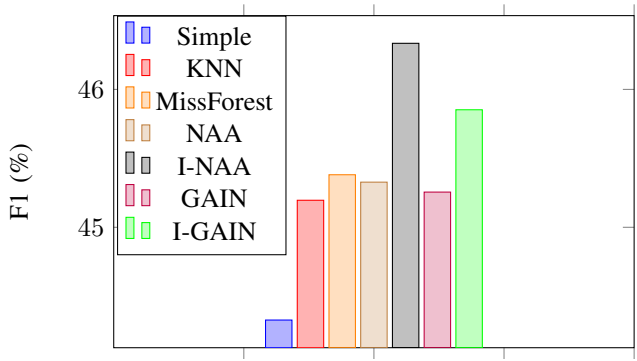
\begin{figure}
\centering
\begin{tikzpicture}

\begin{axis} [ybar,height=6cm,width=8.5cm,xticklabels={,,},ylabel={F1 (\%) },
legend style={at={(0.332,1)}}]

\addplot coordinates {
    (0, 44.327334) 
    
};
\addplot coordinates {
    (0,45.196) 
    
};

\addplot[orange,fill=orange,fill opacity=0.25]  coordinates {
    (0,45.381) 
    
};

\addplot [brown,fill=brown,fill opacity=0.25]coordinates {
    (0,45.32733) 
    
};
\addplot [black,fill=black,fill opacity=0.25]coordinates {
    (0,46.33413) 
    
};
\addplot[purple,fill=purple,fill opacity=0.25]  coordinates {
    (0,45.2552) 
    
};

\addplot[green,fill=green,fill opacity=0.25] coordinates {
    (0,45.85221) 
    
};

\legend{Simple,KNN,MissForest,NAA,I-NAA,GAIN,I-GAIN}
\end{axis}
 
\end{tikzpicture}
\caption{Post-imputation F1-score for different missing value imputation models.}
\label{Fig3}
\end{figure}

\paragraph{\textbf{Results}}
For each dataset produced by a missing value imputation method we perform post-imputation prediction with the results depicted in Fig.~\ref{Fig3}. 

We observe that I-NAA and I-GAIN produce the best post-imputation prediction F1-scores. More specifically, I-NAA is the best performing model outperforming the other approaches by 0.48-2.43\%. We also notice that I-GAIN outperforms GAIN in the post-imputation task too by 0.60\%. Moreover, MissForest produces the third best prediction F1-score accounting for 45.38\%. Finally, we observe that the Simple approach is the worst performing one obtaining an F1-score of 44.10\%. 

\section{Conclusion and Future Work}
In this paper, we studied the case of missing value imputation in a cardiovascular disease dataset. We built upon existing deep learning architectures by introducing improvements fit to the specific case. We evaluated them for both imputation and post-imputation performance. Regarding the task of missing values imputation, we experimented with various missing rates and showed that the introduced approaches outperform the state-of-the-art ones reaching normalised RMSE and AUROC scores up to 0.095, 82.00\% respectively. For the prediction task (post-imputation prediction), findings suggested that our introduced approaches achieved the best imputation F1-scores with a difference of 2.50\% compared to other approaches.

In the future, we aim to evaluate our methods in more medical datasets. Moreover, since these methods are trained and tested on the same dataset, they have been essentially fit to a specific patient distribution. For this reason a model trained on Framingham could be used for missing value imputation on different heart disease data to assess cross-dataset performance. 

\section*{Acknowledgements}

This work was supported by computational time granted from the National Infrastructures for Research and Technology S.A. (GRNET) in the National HPC facility - ARIS - under project ID pa220505.

\bibliographystyle{IEEEtran}
\bibliography{references}

\end{document}